\documentclass[review]{elsarticle}

\usepackage{url}
\usepackage{lineno}
\usepackage{amsmath}
\modulolinenumbers[5]
\usepackage{amssymb}
\usepackage{tabularx}
\usepackage{epsfig}
\usepackage{combelow}
\usepackage{graphicx}

\journal{Journal of Computer Vision and Image Understanding}

\begin{document}

\begin{frontmatter}

\title{Deep-Anomaly: Fully Convolutional Neural Network for Fast Anomaly Detection in Crowded Scenes}

\author[ipm]{M.~Sabokrou\corref{cor1}}
\ead{sabokro@ipm.ir}
\author[sensifie]{M. Fayyaz\corref{cor1}}
\ead{fayyaz@sensifie.com}
\author[iust]{M. Fathy}
\ead{mahfathy@iust.ac.ir}
\author[aut]{Z. Moayed}
\ead{zmoayed@aut.ac.nz}
\author[aut]{and R. Klette}
\ead{rklette@aut.ac.nz}

\cortext[cor1]{Equal Collaboration}
\address[ipm]{School of Computer Science, Institute for Research in Fundamental Sciences (IPM)\\P.o.Box 19395-5746, Tehran, Iran}
\address[sensifie]{Di Hub, Brussels, Belgium}
\address[iust]{Iran University of Science and Technology, Tehran, Iran}
\address[aut]{School of Engineering, Computer and Mathematical Sciences, EEE Department\\
Auckland University of Technology, Auckland, New Zealand }

\begin{abstract}
The detection of abnormal behaviours in crowded scenes has to deal with many challenges. 
This paper presents an efficient method for detection and localization of 
anomalies in videos. Using \textit{fully convolutional neural networks} (FCNs) 
and temporal data, a pre-trained supervised FCN is transferred into an unsupervised FCN
ensuring the detection of (global) anomalies in scenes. High performance in terms 
of speed and accuracy is achieved by investigating the cascaded detection 
as a result of reducing computation complexities. This FCN-based architecture 
addresses two main tasks, feature representation and cascaded outlier 
detection. Experimental results on two benchmarks suggest that detection 
and localization of the proposed method outperforms existing methods 
in terms of accuracy.
\end{abstract}


\end{frontmatter}

\section{Introduction}
\label{sec:Int}

The use of surveillance cameras requires that computer vision technologies need to be 
involved in the analysis of very large volumes of video data. The detection of 
anomalies in captured scenes is one of the applications in this area. 

Anomaly detection and localization is a challenging task in 
video analysis already due to the fact
that the definition of ``anomaly'' is subjective, or context-dependent. 
In general, an event is considered to identify an ``anomaly'' when it occurs 
rarely, or unexpected; for example, see \cite{sabokrou2017}.

Compared to the previously published deep-cascade method in
\cite{sabokrou2017}, this paper proposes and evaluates
a different and new method for anomaly detection. Here we introduce and study a 
modified pre-trained {\it convolutional neural network} (CNN) 
for detecting and localizing anomalies.  
In difference to~\cite{sabokrou2017}, the considered CNN is not 
trained from scratch but ``just'' fine-tuned. More in detail, for processing a video frame
\cite{sabokrou2017} outlined a method where the frame 
was first divided into a set of patches, then the anomaly detection 
was organised based on levels of patches.  In difference to that,
the input of the proposed CNN algorithm is a full video frame in this paper. 
As a brief preview, the new method is methodically simpler 
but faster in both the training and testing phase where the accuracy 
of anomaly detection is comparable to the accuracy 
of the method presented in~\cite{sabokrou2017}. 

In the context of crowd scene videos, anomalies are formed 
by rare shapes or rare motions. 
Due to the fact that looking for unknown shapes 
or motions is a time-consuming task, state-of-the-art 
approaches learn regions or patches of normal frames as reference models. 
Indeed, these reference models include normal motion or shapes 
of every region of the training data. In the testing phase, those regions 
which differ from the \textit{normal model} are considered to be abnormal. 
Classifying these regions into normal and abnormal requires extensive 
sets of training samples in order to describe the properties of 
each region efficiently.

There are numerous ways to describe region properties. Trajectory-based 
methods have been used to define behaviours of objects. Recently, 
for modeling spatio-temporal properties of video data, low-level features 
such as {\it histogram of gradients} (HoG) and 
{\it histogram of optic flow} (HoF) are used. These trajectory-based methods 
have two main disadvantages. They cannot handle occlusion problems,
and they also suffer from high complexity, especially in crowded scenes.

CNNs proved recently to be useful for defining effective
data analysis techniques for various applications. CNN-based approaches 
outperformed state-of-the-art methods in different areas including 
image classification \cite{alexnet}, object detection \cite{girshik2014}, or 
activity recognition \cite{Simonyan2014}. It is argued that 
handcrafted features cannot efficiently represent normal 
videos \cite{sabokrou2015,xu2015,sabokrou2016}. In spite of these benefits, 
CNNs are computationally slow, especially when considering 
block-wise methods \cite{girshik2014, giusti2013}. Thus, dividing 
a video into a set of patches and representing them by using CNNs, 
should be followed by a further analysis about possible ways of speed-ups. 

Major problems in anomaly detection using CNNs are as follows:
\begin{enumerate}
    \item Too slow for patch-based methods; thus, CNN is considered 
    as being a time-consuming procedure.
    \item Training a CNN is totally supervised learning; thus, the detection
    of anomalies in real-world videos suffers from a basic
    impossibility of training large sets of samples from non-existing classes of anomalies.
\end{enumerate}
    
Due to these difficulties, there is a recent trend to optimize 
CNN-based algorithms in order to be applicable in practice. 
Faster-RCNN \cite{ren2015} takes advantage of convolutional 
layers to have a feature map of every region in the input data, 
in order to detect the objects.
For semantic segmentation, methods such 
as \cite{Shelhamer2016, long2015} use {\it fully convolutional networks} (FCNs)
for traditional CNNs to extract regional features. Making traditional 
classification CNNs to work as a fully convolutional network 
and using a regional feature extractor reduces 
computation costs. 
In general, as CNNs or FCNs are supervised methods, 
neither CNNs nor FCNs are capable for solving anomaly detection tasks,  

To overcome aforementioned problems, we propose 
a new FCN-based structure to extract the 
distinctive features of video regions. This new approach includes 
several initial convolutional layers of a pre-trained CNN 
using an AlexNet model \cite{alexnet} and an additional 
convolutional layer. AlexNet, similar to \cite{zhou2014}, is a 
pre-trained model proposed for image classification by 
using ImageNet \cite{imageNet1,imagenet} and the 
MIT places dataset \cite{mit}. 
Extracted features, by following this approach, are sufficiently
discriminative for anomaly detection in video data. 

In general, entire frames are fed to the proposed FCN. 
As a result, features of all regions are extracted efficiently. 
By analysing the output, anomalies in the video are extracted and localized. 
The processes of convolution and pooling, in all of the CNN layers, run concurrently.
A standard NVIDIA TITAN GPU processes $\approx 370$ 
{\it frames per second} (fps) when analyzing (low-resolution) 
frames of size $320 \times 240$. This is considered to be ``very fast''.

Convolution and pooling operations in CNNs are responsible for 
extracting regions from input data using a specific stride and size. 
These patch-based operations provide a description 
for each extracted region. Detected features in the output and the 
corresponding descriptors distinguish a potential region in a 
set of video frames. Both convolution and pooling operations are invertible.
However, a roll-back operation generates a receptive field (a region in a frame) 
from deeper layers to more shallow layers of the network. This 
receptive field results in the generation of feature vectors.

In this paper, we propose a method for detecting and localizing abnormal 
regions in a frame by analyzing the output of deep layers in an FCN.
The idea of localizing a receptive field is inspired by the
faster-RCNN in \cite{ren2015}, and OverFeat in \cite{Overfeat, Simonyan2014}. 

This paper uses the structure of a CNN for patch-based operations 
in order to extract and represent all patches in a set of frames.
A generated feature vector, while using the CNN for each detected region, 
is fitted to the given image classification task. 

Similar to~\cite{oquab2014}, we use a transfer learning method to 
gain a better description for each region. We evaluate our method for finding 
the best intermediate convolutional layer of the CNN. Then, a new 
convolutional layer is added after
the best-performing layer of the CNN. 
The kernels of a pre-trained CNN are  
adjusted based on pre-training, and 
considered to be constant in our FCN; the parameters of the final new convolutional
layer are trained based on our training frames.

In other words, all regions generated by the pre-trained CNN are 
represented by a sparse-auto-encoder as a feature vector of 
length $h$ which is the 
hidden size of the auto-encoder.
We find that the feature set, generated by a
pre-trained CNN, is sufficiently discriminative for modeling 
``many''  regions. To make the process more accurate, those regions which are 
classified with low confidence, are given to the final convolutional layer 
for further representation and classification. 

In fact, two 
Gaussian models are defined based on the description of all normal 
training regions. The first model is 
generated by the $k^{th}$ layer of the CNN, while the second model is
based on its transformation by the 
$(k+1)^{th}$ convolutional layer.

In the testing phase, those regions which differ significantly
from the first Gaussian model, are labeled as being a {\it confident anomaly}.
Those regions which fit completely to the first model are 
labeled as being {\it normal}. The rest of the regions, being by a minor difference below the threshold,
are represented by a sparse-auto-encoder and evaluated more carefully 
by the second Gaussian model. 
This approach is similar to a cascade classifier defined by two stages; it is 
explained in the next sections.   

The main contributions of this paper are as follows:
\begin{itemize}
\item To the best of our knowledge, this is the first time that 
an FCN is used for anomaly detection.  
\item We adapt a pre-trained classification CNN to an FCN for generating video regions to describe motion and shape concurrently.
\item We propose a new FCN architecture for time-efficient 
anomaly detection and localization.
\item The proposed method performs as well as state-of-the-art methods, but our 
method outperforms those with respect to time; we have real-time for typical applications.  
\item We  achieved a processing speed of 370 fps on a standard GPU; this
is about three times faster than the fastest existing method reported so far.
 \end{itemize}
 
 Section~\ref{sec:rw} provides a brief survey on existing work. We 
 present the proposed method
in Section~\ref{sec:ps} including the overall scheme of our method,  
and also details for anomaly detection and localization, and for the evaluation of 
different layers of the CNN for performance optimization. 
Qualitative and quantitative experiments are described in Section~\ref{sec:er}. 
Section~\ref{sec:c} concludes the paper.

\section{Related Work}
\label{sec:rw}

Object trajectory estimation is often of interest in cases of 
anomaly detection; see \cite{Jiang2011,wu2010,Piciarelli2006,Piciarelli2008,Antonakaki2009,Calderara2011, morris2011,hu2006,Tung2011}. 
An object shows an anomaly if it does not follow learned normal trajectories. 
This approach usually suffers from many weaknesses, such as 
disability to efficiently handle occlusions, and being
too complex for processing crowded scenes.

To avoid these two weaknesses, it is proposed to use spatio-temporal low level features
such as optical flow or gradients. 
Zhang et al. \cite{zhang2005} use a \textit{Markov random field} (MRF) to model 
the normal patterns of a video with respect to a number of features, 
such as rarity, unexpectedness, and relevance. 
Boiman and Irani~\cite{Boiman2007} consider an event as being abnormal if 
its reconstruction 
is impossible by using previous observations only.
Adam et al.~\cite{adam2008} use an exponential distribution for modeling the 
histograms of optical flow in local regions.

A {\it mixture of dynamic textures} (MDT) 
is proposed by Mahadevan et al.~\cite{Mahadevan2010} for representing 
a video. In this method, the represented features fit into a Gaussian mixture model. 
In \cite{Li2014}, 
the MDT is extended and explained in more details.
Kim and Grauman~\cite{kim2009} exploit 
a {\it mixture of probabilistic PCA} (MPPCA) 
model for representing local optical flow patterns. They also use 
an MRF for learning  the normal patterns. 

A method based on motion properties of pixels for 
behavior modeling is proposed 
by Benezeth et al.~ \cite{Benezeth2009}. They described 
the video by learning a 
co-occurrence matrix for normal events across space-time.  
In~\cite{kratz2009}, a Gaussian model is fitted into spatio-temporal 
gradient features, and a {\it hidden Markov model} (HMM) is used for 
detecting the abnormal events. 

Mehran et al.~\cite{mehran2009} introduce {\it social force} (SF)
as an efficient technique for abnormal motion modeling of crowds. 
Detection of abnormal behaviors using a method based on spatial-temporal 
oriented energy filtering in proposed by In~\cite{Zaharescu2010}.

Cong et al.~\cite{cong2011} construct an over-complete 
normal basis set from normal data. A patch is considered to 
be abnormal if reconstructing it with this basis set is
not possible. 

In \cite{antic2011}, a scene parsing approach is proposed by Antic et al. All object hypotheses 
for the foreground of a frame are explained by normal training.  
Those hypotheses, which cannot be explained by normal training, are considered to show anomaly.
Saligrama et al. in~ \cite{saligrama2012} propose a method based on the clustering of the test data 
using optic-flow features. Ullah et al. \cite{ullah2012} introduced 
an approach based on a cut/max-flow algorithm for segmenting the crowd motion.  
If a flow does not follow the regular motion model, it is considered as being an anomaly.
Lu et al.~\cite{lu2013} propose a fast (140-150 fps) anomaly detection method based on sparse representation. 

In~\cite{Roshtkhari2013}, an extension of the 
{\it bag of video words} (BOV) approach is used by  Roshtkhari et al. 
A context-aware anomaly detection algorithm is proposed in~\cite{zhu2013}, where the 
authors represent the video using motions and the context of videos. 
In~\cite{cong2013}, a method for modeling both motion and shape with 
respect to a descriptor (named ``motion context'')  is proposed;
they consider  anomaly detection as a matching problem.  
Roshkhari et al.~\cite{Roshtkhari2013sec} introduce a method for learning 
the events of a video by using the construction of a hierarchical codebook 
for dominant events in a video. Ullah~et al.~\cite{ullah2013}  
learn an MLP neural network using trained particles to extract the video 
behavior. A {\it Gaussian mixture model} (GMM) is exploited for learning the behavior of 
particles using extracted features.
In addition, in~\cite{ullah2014}, an MLP neural network for extracting 
the corner features from normal training samples is proposed; authors also 
label the test samples using that MLP. 

Authors of~\cite{Ullah2014Sec} extract  
corner features and analyze them based on their properties of 
motion by an enthalpy model, a random forest with corner features 
for detecting abnormal samples. 
Xu et al.~\cite{Xu2014} propose a unified anomaly energy function 
based on a hierarchical activity-pattern 
discovery for detecting anomalies.

Work reported in \cite{sabokrou2015,xu2015} models
normal events based on a set of representative features which are 
learned on auto-encoders \cite{Vincent2008}. They use a one-class classifier 
for detecting anomalies as being outliers compared to the target (normal) class. 
See also the beginning of Section~\ref{sec:Int} where we briefly
reviewed work reported in \cite{sabokrou2017}; this paper proposes
a cascaded classifier which takes advantage of two deep neural networks
for anomaly detection. Here, challenging patches are identified at first by 
using a small deep network; then the neighboring patches are passed 
into another deep network for further classification.

In~\cite{Mousavi2015}, the {\it histogram of oriented tracklets} (HOT) is 
used for video representation and anomaly detection.  
A new strategy for improving HOT is also introduced in this paper.   
Yuan et al.~\cite{yuan2015} propose an informative {\it structural context descriptor}
(SCD) to represent a crowd individually. 
In this work, a (spatial-temporal) SCD variation of a crowd is 
analyzed to localize the anomaly region. 

An unsupervised deep learning approach is used in \cite{feng2017learning}
for extracting anomalies in crowded scenes. In this approach, 
shapes and features are extracted using a PCANet \cite{PCAnet} 
from 3D gradients. Then, a deep {\it Gaussian mixture model} (GMM) is 
used to build a model that defines the event patterns. A
PCANet is also used in \cite{Zhijun}. In this study, authors 
exploit the {\it human visual system} (HVS) to define features in 
the spatial domain. On the other hand, a {\it multi-scale histogram of optical flow} 
(MHOF) is used to represent motion features of the video. 
PCANet is adopted to exploit these spatio-temporal features 
in order to distinguish abnormal events.

A hierarchical framework for local and global anomaly detection 
is proposed in \cite{cheng2015}. Normal interactions are 
extracted by finding frequent geometric relationships between sparse 
interest points; authors model the normal interaction template by 
Gaussian process regression. Xiao et al.~\cite{xiao2015} 
exploit {\it sparse semi-nonnegative matrix factorization} (SSMF) for 
learning the local pattern of pixels. Their method learns a probability model 
by using local patterns of pixels for considering both the spatial and 
temporal context. Their method is totally unsupervised. Anomalies are detected by  
the learned model. 

In~\cite{lee2015}, an efficient method for representing 
human activities in video data with respect to motion characteristics is introduced 
and named as {\it motion influence map}. Those blocks of a frame 
which have a low occurrence are labelled as being abnormal.
A spatio-temporal CNN is developed in \cite{Zhou2016} to define anomalies in 
crowded scenes; this CNN model is designed to detect features in 
both spatial and temporal dimensions using spatio-temporal convolutions.

Li et al. \cite{Li2015} propose an unsupervised framework for detecting 
the anomalies based 
on learning global activity patterns and local salient behavior patterns 
via clustering and sparse coding. 

\section{Proposed Method}
\label{sec:ps}

This section explains at first the overall outline of the method. Then, 
a detailed description of the proposed method is given.

\subsection{Overall Scheme}

Abnormal events in video data are defined in terms of irregular 
shapes or motion, or possibly a combination of both. As a result 
of this definition, identifying the shapes and motion is an essential 
task for anomaly detection and localization.
In order to identify the motion properties of events, we need a series of frames. 
In other words, a single frame does not include motion properties; 
it only provides shape information of that specific frame. 

For analyzing both shape and motion, we 
consider the pixel-wise {\it average} 
of frame $I_{t}$ and previous frame $I_{t-1}$, denoted by $I^\prime_{t}$ (not to be confused 
with a derivative),
\begin{equation}
I^\prime_{t}(p)=\frac{I_{t}(p)+I_{t-1}(p)}{2}
\label{eq:av}
\end{equation}
where $I_{t}$ is $t^{th}$ frame in the video. 
For detecting 
anomalies in $I_{t}$, we use the sequence 
$D_t=\langle I'_{t-4},I'_{t-2},I'_{t}\rangle$.

We start with this sequence $D_t$
when representing video frames on grids of decreasing size $w \times h$. 
$D_t$ is defined on a grid $\Omega_0$ of size $w_0 \times h_0$.
The sequence $D_t$ 
is subsequently passed on to an FCN, defined 
by the $k^{th}$ intermediate convolutional layer, for $k=0, 1, \ldots, L$, 
each defined on a grid $\Omega_k$ of size $w_k \times h_k$,
where $w_k > w_{k+1}$, and $h_k > h_{k+1}$. We use $L=3$ for 
the number of convolutional layers.

The output of the $k^{th}$ intermediate convolutional layer of the FCN are   
feature vectors 
$f_k \in \mathbb{R}^{m_k}$ (i.e. each containing $m_k$ real feature values),
satisfying $m_k \le m_{k+1}$, starting with 
$m_0=1$.
For the input sequence $D_t$, the output of the $k^{th}$ convolutional layer 
is a matrix of vector values:
\begin{equation}
\left\{f_{k}^t(i,j,1:m_k)\right\}_{(i,j)=(1,1)}^{\;(w_k,h_k)} 
= \left\{\left[f_{k}^t(i,j,1),\ldots,f_{k}^t(i,j,m_k)\right]^\top\right\}_{(i,j)=(1,1)}^{\;(w_k,h_k)}
\label{eq:p}
\end{equation}
Each feature vector $f_k^t(i,j,1:m_k)$ is derived from a specific {\it receptive field}
(i.e. a sub-region of input $D_t$). 

In other words, first, a high-level description of $D_t$ is provided
for the $t^{th}$ frame of the video. Second, $D_t$ is 
represented subsequently by the $k^{th}$
intermediate convolutional layer of the FCN, for $k=1, \ldots, L$. 
This representation is used for identifying a set of partially pairwise
overlapping regions in $\Omega_k$, called the {\it receptive fields}.
Hence, we represent frame $I_{t}$ at first by sequence $D_t$
on $\Omega_0$, and then by $m_k$ maps 
\begin{equation}
f_{k,l}=\left\{f_{k}^t(i,j,l)\right\}_{(i,j)=(1,1)}^{\;(w_k,h_k)}, \; \mbox{for} \; l=1,2,\ldots,m_k
\end{equation}
on $\Omega_k$, for $k=1, \ldots, L$. Recall that
the size $w_k \times h_k$ decreases with increases of $k$ values.

Suppose that we have $q$ training frames from a video which are considered to be normal. 
To represent these normal frames with respect to the $k^{th}$ convolutional 
layer of the FCN (AlexNet without its fully connected layers), we 
have $w_k \times h_k \times q$ vectors of length $m_k$,
defining our 2D normal region descriptions; they are generated
automatically by a pre-trained FCN .
For modeling the normal behavior, a Gaussian distribution is fitted
as a one-class classifier to the descriptions of normal regions so that it 
defines our {\it normal reference model}. 
In the testing phase, a test frame $I_{t}$ is described in a similar 
way by a set of regional features. Those regions which differ from the 
normal reference model are labeled as being {\it abnormal}.
In particular, the features generated by a pre-trained CNN 
($2^{nd}$ layer of AlexNet)  are sufficiently discriminative.
These features are learned based on a set of independent images which are not necessarily 
related to video surveillance applications only. 

Consequently,  suspicious regions are represented by a ``more discriminant'' 
feature set. This new representation leads to a better performance for 
distinguishing abnormal regions from normal ones. In other words, we transform the generated features 
by AlexNet into an anomaly detection problem. This work is done by an 
auto-encoder which is trained on all normal regions. 
As a result, those suspicious $f_k^t(i,j,1:m_k)$ regions are passed to 
an auto-encoder to have a better representation. 
This is done by the 
$(k+1)^{st}$ convolutional layer whose kernels are learned by a 
sparse auto-encoder.

\begin{figure}[t!]
\center
  \includegraphics[width=1\linewidth]{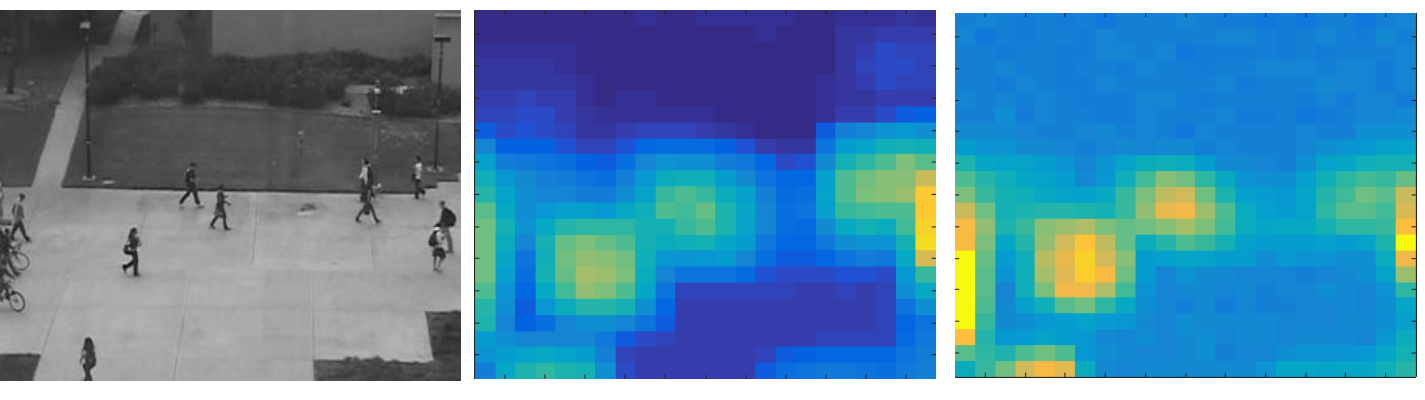}
   \caption
   {Effect of representing receptive fields with an
   added convolutional layer.
      {\it Left:} Input frame. {\it Middle:} 
   Heat-map visualisation of the $2^{nd}$ layer of a pre-trained FCN.  
   {\it Right:}  Heat-map visualisation of the $3^{nd}$ layer of a pre-trained FCN with 
   added convolutional layer.
   }
\label{fig:heatmap}
\vspace{-2mm}
\end{figure}

Let $T_k^t(i,j,1:m_k)$ be 
the transformed representation of $f_k^t(i,j,1:m_k)$  by a sparse auto-encoder;
see Figure~\ref{fig:heatmap}. The abnormal region is visually more distinguishable in the
heat-map when the regional descriptors are represented again by the 
auto-encoder (i.e. the final convolutional layer).

Then, for the new feature space, those regions  
which differ from the normal reference model are labeled as being abnormal. 
This proposed approach ensures both accuracy and speed. 

\begin{figure}[b!]
\center
  \includegraphics[width=.95\linewidth]{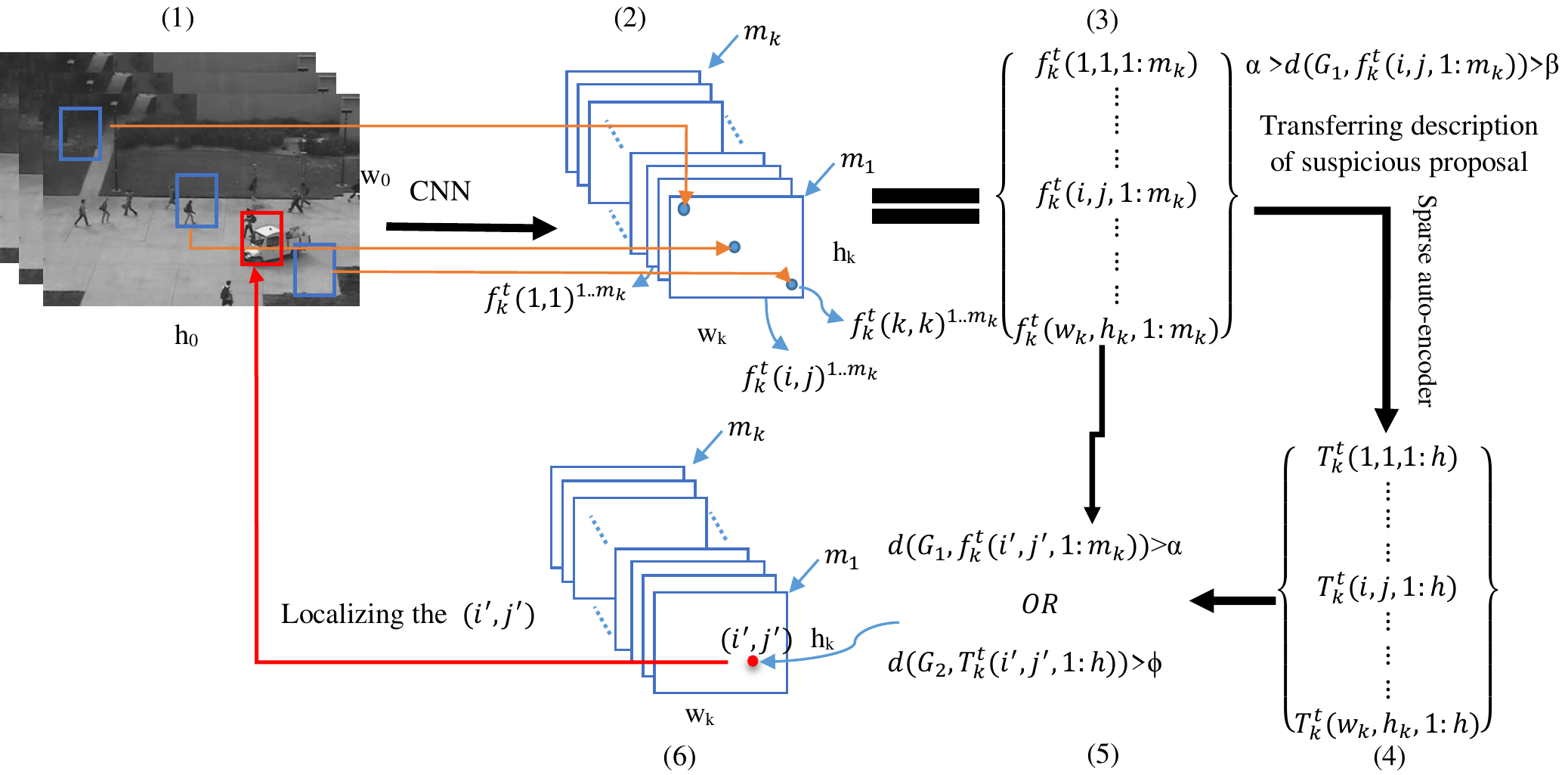}
   \caption
   {Schematic sketch of the proposed method. 
   (1) Input video frame of size $w_0 \times h_0$. 
   (2,3) Description of regions of size $h_k\times w_k$ 
   generated by the $k^{th}$ layer of the FCN. 
   (4) Transformed feature 
   domain using a sparse auto-encoder. 
   (5) Joint anomaly detector. 
   (6) Location of descriptions which identify anomalies. 
   }
\label{fig:os}
\vspace{-2mm}
\end{figure}

Suppose that $f(i,j,1:m_k) \in \mathbb{R}^{m_k}$  
is the description of an abnormal region. By 
moving backward from the $k^{th}$ to the $1^{st}$ layer of the FCN, 
we can identify regions in input frames with descriptions $f_k^t(i,j,1:m_k)$. 
This is due to the fact that convolution and
mean pooling operator of the FCN (from $1^{st}$ to $2^{nd}$ layer) 
are approximately invertible. 

For instance, the $1^{st}$ and $2^{nd}$ convolutional layer, and the $1^{st}$ 
sub-sampling layer are called C$_{1}$, C$_{2}$, and $S_{1}$, respectively.
As usual, $(.)^{-1}$ identifies below the inverse of a function. The exact location of  
description $f_k^t(i,j,1:m_k)$ in the $D_t$ sequence (the input of the FCN)
is located at $C_{1}^{-1}(S_1^{-1}(C_{2}^{-1}(f(i,j,1:m_k)))$. See the following sections
for more details.

Figure~\ref{fig:os} shows the work-flow of the proposed detection method. 
First, input frames are passed on to a pre-trained FCN. 
Then, $h_k\times w_k$  regional feature vectors are generated in the output 
of the $k^{th}$ layer. These feature vectors are verified using Gaussian classifier $G_1$.
Those patches, which differ significantly from $G_1$ as a normal reference model, 
are labeled as being abnormal. 
More specifically, $G_1$ is a Gaussian distribution which is fitted to all of the 
normal extracted regional feature vectors; 
regions which completely differ from $G_1$ are considered to be an anomaly. 

Those suspicious regions which are fitted with low confidence are given 
to a sparse auto-encoder. At this stage, we also label these regions based on 
Gaussian classifier $G_2$ which works similar to $G_1$.   
$G_2$ is also a Gaussian classifier, trained on all extracted regional feature vectors from 
training video data which are represented by an auto-encoder.  
Finally, the location of those abnormal regions can be annotated by a roll-back on the FCN.

\subsection{Anomaly Detection}
\label{sec:ad}

In this paper, the video is represented using a set of regional features. 
These features are extracted densely and their description is given by 
feature vectors in the output of the $k^{th}$ convolutional layer.
See Equ.~(\ref{eq:p}). 

Gaussian classifier $G_1(.)$ is fitted to all normal regional features generated by the FCN. 
Those regional features for which their distance to $G_1(.)$ is bigger than threshold 
$\alpha$ are considered to be abnormal.
Those ones that are compatible to $G_1$ (i.e. their distance is 
less than threshold $\beta$) are labeled as being normal. 
A region is {\it suspicious} if it has a distance to $G_1$ being
between $\alpha$  and $\beta$. 

All suspicious regions are given to the next convolutional layer 
which is trained on all normal regions generated by the pre-trained FCN. 
The new representation of these suspicious regions is more discriminative 
and denoted by
\begin{equation}
T_{k,n}=\left\{T_{k}^t(i,j,n)\right\}_{(i,j)=(1,1)}^{\;(w'_k,h'_k)}, \;\; \mbox{for} \; n=1,2,\ldots,h
\end{equation}  
where $h$ is the size of the feature vectors generated by the auto-encoder, 
which equals the size of the hidden layers. 

In this step, 
only the suspicious regions are processed. Thus, some points
$(i,j)$ in grid $(w_k,h_k)$ are ignored and not analysed in the grid $(w'_k,h'_k)$. 
Similar to $G_1$,
we create a Gaussian classifier $G_2$ on all of the normal training regional features 
which are represented by our auto-encoder. Those regions which are not sufficiently 
fitted to $G_2$ are considered to be abnormal. 

Equations~(\ref{eq:a}) and (\ref{eq:a1}) summarize anomaly detection 
by using two fitted Gaussian classifiers. First, we have that
\begin{equation}
G_1(f_k^t(i,j,1:m_k))=
\begin{cases}
{\rm Normal} &    \mbox{if} \;\; d(G_1, f_k^t(i,j,1:m_k)) \leq {\rm \beta} \\
{\rm Suspicious } &   \mbox{if} \;\; \beta <  d(G_1, f_k^t(i,j,1:m_k)) < {\rm  \alpha}\\
{\rm Abnormal} &   \mbox{if} \;\; d(G_1, f_k^t(i,j,1:m_k)) \geq {\rm \alpha} \\
\end{cases}  \\  
\label{eq:a}
\end{equation}
Then, for a suspicious region represented by $T_k^t(i,j,1:h)$, we have that:
\begin{equation}
G_2(T_k^t(i,j,1:h_k))=
\begin{cases}
{\rm Abnormal} &    \mbox{if} \;\; d(G_2, T_k^t(i,j,1:h)) \geq {\rm \phi} \\
{\rm Normal} &   \mbox{otherwise} \\
\end{cases}
\label{eq:a1}
\end{equation}
Here, $d(G,{\bf x})$ is the Mahalanobis distance of a
regional feature vector {\bf x} from the $G$-model. 

\subsection{Localization}
\label{sec:lo}

The first convolutional layer has $m_1$ 
kernels of size $x_1\times y_1$.  
They are convolved on sequence $D_t$ for 
considering the $t^{th}$ frame. As a result of this convolution, 
a feature is extracted. 

Recall that each  region for the input of the FCN 
is described by a feature vector of length $m_1$. 
In this continuous process, we have $m_k$ maps as output 
for the k$^{th}$ layer.
Consequently, a point in the output of the $k^{th}$ layer 
is a description for a subset of overlapping $(x_1\times y_1)^{th}$ receptive fields 
in the input of the FCN.

The order of layers in the modified version of AlexNet is 
denoted by 
\begin{equation}
\text{AlexNet Order} \rightarrow [C_1, S_1, C_2,S_2, C_3, fc_1, fc_2]
\end{equation}
where $C$ and $S$ are a convolutional layer and a sub-sampling layer, respectively. 
The two final layers are fully connected.

Assume that $n$ regional feature vectors $(i_1,j_1) \cdots (i_n,j_n)$, generated in layer $C_k$ on 
grid $\Omega_k$, are identified as showing an anomaly.
The location $(i,j)$ in $\Omega_k$ corresponds to
\begin{equation}
C_{1}^{-1}(\cdots S_{k-1}^{-1}(C_{k}^{-1}(i,j)))
\end{equation}
as the rectangular region in the original frame. 

Suppose we 
have $m_k$ kernels 
of size $x_k \times y_k$ which are then convolved with stride $d$ on the output 
of the previous layer of $C_k$.  
$C^{-1}_k(i,j)$ is the (rectangular) 
set of all locations in $\Omega_{k-1}$ which are mapped in the FCN on $(i,j)$ in $\Omega_k$.
Function $S^{-1}_k$ is defined in an analogous way. 

The sub-sampling (mean pooling) layer can also be considered as
a convolutional layer which has only one kernel. 
Any region, detected as being an abnormal region in the original frame (i.e. in $\Omega_0$),  
is then a combination of some overlapping and large patches.
This leads to a poor localization performance. 

As a case in point, 
a detection in the $2^{nd}$ layer causes 
$51 \times 51$ overlapping receptive fields. To 
achieve more accuracy in anomaly detection,
those pixels in $\Omega_0$ are identified to show an anomaly which are 
covered by more than $\zeta$ related receptive 
fields (we decided for $\zeta$=3 experimentally).   

\subsection{FCN Structure for Anomaly Detection}

This section analyses the quality of different layers of a
pre-trained CNN for generating regional feature vectors. 
We adapt (in this paper in general)
a classification by CNN into an FCN by solely using convolutional layers. 
Selecting the best layer for
representing the video is crucial considering the following two 
aspects: 
\begin{itemize}
\item[(1)] Although deeper features are usually more discriminative, 
using these deeper features is time-consuming. In addition, since the 
CNN is trained for image classification, 
going deeper may create over-fitted features for image classification. 
\item[(2)] Going deeper leads to larger receptive fields in the input data; 
as a result, the likelihood of inaccurate localization increases which 
then has inverse effects on performance.
\end{itemize}
For the first two convolutional layers of 
our FCN model, we use a modified version of AlexNet named 
{\it Caffe reference model}.\footnote
      {
      Caffe is a framework maintained 
      by UC Berkeley \cite{Jia2014}.
      }

This model is trained on 1183 categories, each with 205 scene categories 
from the MIT places database \cite{mit}, and 978 object categories 
from the train data of ILSVRC2012 (ImageNet) \cite{imageNet1, imagenet} 
having 3.6 million images. 

The implemented FCN has three convolutional layers.  
For finding the best convolutional layer $k$,
we set initially $k$ to 1, and then increase it to 3.   
When the best $k$ is decided, deeper layers are ignored. 

The general findings are described at an abstract level.
First we use the output of layer $C_1$. For distinguishing 
abnormal from normal regions, corresponding receptive fields are 
small in size, and generated features are not capable of 
achieving the suitable results. Therefore, here we have lots of false positives. 
Later, the output of $C_2$ is used as a deeper layer.  
At this stage, we achieve better performance compared to 
$C_1$ due to the following reasons: 
A corresponding receptive field in the input frames of $C_1$ 
is now sufficiently large, 
and the deeper features are more discriminative. 

At $k=3$, we have the results in layer $C_3$ as output.
Although the capacity of the network increases, results are not as good  as 
for the $2^{nd}$ convolutional layer. It seems that by adding one more layer,
we achieved deeper features; however, these features are also likely to over-fit the 
image classification tasks since the network is trained for ImageNet. 

\begin{figure}[t!]
\vspace{-3mm}
\center
  \includegraphics[width=1\linewidth]{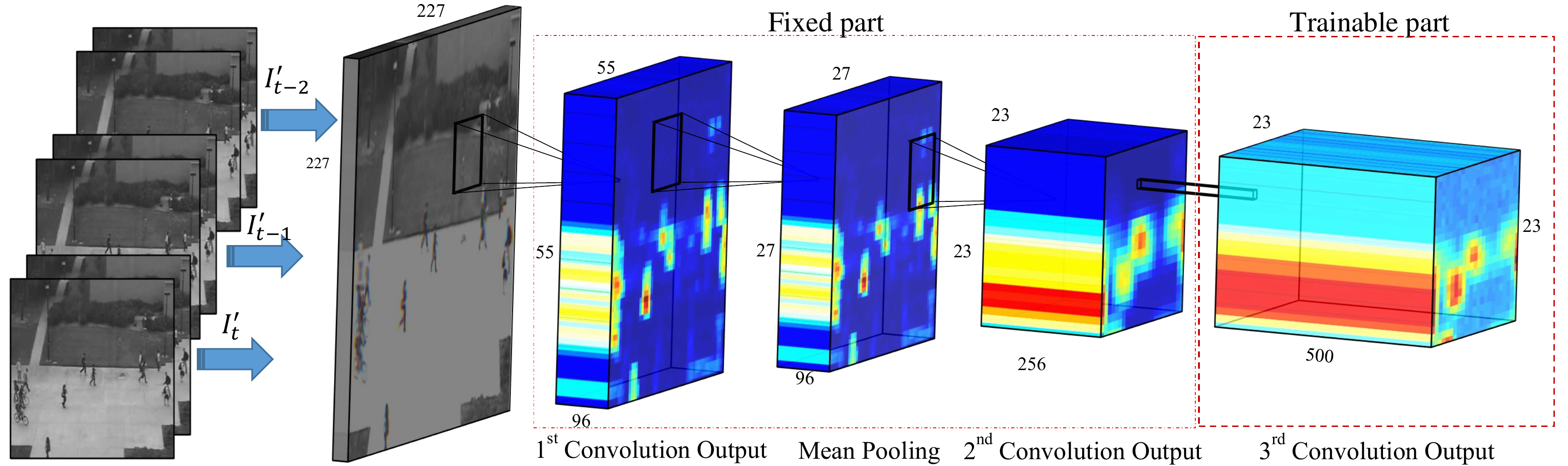}
\vspace{-3mm}
   \caption
      {   
Proposed FCN structure for detecting anomalies.  
This FCN is only used for regional feature extraction. At later stages, 
two Gaussian classifiers are embedded for labeling abnormal regions.
  }
\label{fig:arch}
\end{figure}

Consequently, we decided for the $C_2$ output for extracting regional features. 
Similar to \cite{oquab2014}, we transformed the description of each generated 
regional feature using a convolutional layer; the
kernels of the layer are learned using a sparse auto-encoder.
This new layer is called $C_T$ that is on top of the $C_2$ layer of the CNN.  
The combination of three 
(initial) layers of a pre-trained CNN (i.e. $C_1$, $S_1$, and $ C_2$) 
with an additional (new) convolutional  layer is our new architecture 
for detecting anomalies. Figure~\ref{fig:arch} shows the proposed FCN structure.
To emphasise further the effects of using this structure, see
Tables~\ref{tab:l} to \ref{tab:3}. 

Table~\ref{tab:l} shows the performance 
of different layers of the pre-trained CNN. 

\begin{table}[h!]
\caption{Evaluating CNN convolutional layers for anomaly detection}
\begin{center}
\begin{tabular}{lccccc}
\hline
Layer   & Output in $C_1$\qquad & \qquad Output in $C_2$ \qquad & \qquad Output in $C_3$  \\
\hline
Proposed size & $11 \times 11$ & $51 \times 51$ & $67 \times 67$  \\
Frame-level EER &40\% & 13\% & 20\%\\
 Pixel-level EER & 47\% & 19\% & 25\%\\
\hline
\end{tabular}
\end{center}
\label{tab:l}
\end{table}

\begin{table}[b!]
\vspace{-3mm}
\caption{Effect of the number of kernels in the $(k+1)^{th}$ convolutional layer, 
used for representing regional features when using $C_2$ as outputs}
\vspace{-3mm}
\begin{center}
\begin{tabular}{lcccc}
\hline
Number of kernels   & \qquad 100 \qquad& \qquad 256 \qquad & \qquad 500  \qquad & \qquad  500 \& two  classifiers\\
\hline
 Frame-level EER  & \qquad 19\% &\qquad 17\% & \qquad 15\%& \qquad \textbf{11\%}\\
\hline
\end{tabular}
\end{center}
\label{tab:2}
\end{table}

Table~\ref{tab:2} reports the performance of using the proposed architecture 
with different numbers of kernels in the $(k+1)^{th}$ convolutional layer.
We represent video frames with our FCN. A Gaussian classifier
is exploited at the final stage of the FCN (see the performance 
for 100, 256, and 500 kernels in Table~\ref{tab:2}). We also evaluated the performance
when two Gaussian classifiers are used in a  
similar approach to a cascade.  
The frame-level and pixel-level EER measures 
are introduced in the next section. Recall that the smaller values of EER, the ``better'' performance.

Table~\ref{tab:3} reports the performance of processing the network outputs in $C_2$ output and $C_T$ output with cascaded classifiers.
 
\begin{table}[ht]
	\caption{Effect of adding the 
	$(k+1){th}$ convolutional layer, used for representing regional features when using $C_2$ for outputs}
	\begin{center}
		\begin{tabular}{cccccc}
			\hline
			Layer   & \qquad $C_2$ \qquad & \qquad$C_T$ and two classifiers\\
			\hline
			Frame-level EER  & \qquad 13\%&\textbf{11\%}\\
			\hline
		\end{tabular}
	\end{center}
	\label{tab:3}
	\vspace{-3mm}
\end{table}

Our results when evaluating the different CNNs confirm that 
the proposed CNN architecture
is the best architecture 
for the studied data.

\section{Experimental Results}
\label{sec:er}

We evaluate the performance of the proposed method on 
UCSD~\cite{UCSD} and Subway benchmarks~\cite{adam2008}. We show that
 our proposed method detects anomalies at high speed,
 similar to a real-time method in video surveillance, with equal or even better performance
 than other state-of-the-art methods. 
 
 For implementing our deep-anomaly architecture we use the Caffe library \cite{Jia2014}. 
 All experiments are done using a standard \textit{NVIDIA TITAN GPU} 
 with \textit{MATLAB 2014a}.

\subsection{UCSD and Subway Datasets}

To evaluate and compare our experimental results, we use two datasets.

\vspace{2mm}
\textit{\textbf{UCSD Ped2}~\cite{UCSD}}.
Dominant dynamic objects in this dataset are walkers where
crowd density varies from low to high.
An appearing object such as a car, skateboarder, wheelchair, or bicycle is 
considered to create an anomaly. All training frames in this dataset 
are normal and contain pedestrians only. 
This dataset has 12 sequences for testing, and 16 video sequences for training,
with $320 \times 240$  resolution. For evaluating the localization, the ground truth 
of all test frames is available. The total numbers of abnormal  
and normal frames are $\approx$2,384 and $\approx$2,566, respectively.

\vspace{2mm}
\textit{\textbf{Subway}~\cite{adam2008}}.
This dataset contains two
sequences recorded at the entrance (1 h and 36 min,
144,249 frames) and exit (43 min, 64,900 frames) of a subway station.   
People entering and exiting the station usually behave normally. 
Abnormal events are defined by people moving
in the wrong direction (i.e. exiting the entrance or entering the
exit), or avoiding payment. This dataset has two limitations:
The number of anomalies is low, and there are
predictable spatial localizations (at entrance or exit regions).

\subsection{Evaluation Methodology}
\label{sec:em}

We compare our results with state-of-the-art methods 
using a {\it receiver operating characteristic} (ROC) curve,  
the {\it equal error rate} (EER), and the {\it area under curve} (AUC). 
Two measures at frame level and pixel level are used, which 
are introduced in \cite{Mahadevan2010} and often exploited in
later work. According to these measures, frames are 
considered to be abnormal (positive) or normal (negative). 
These measures are defined as follows:
%
\begin{itemize}
\item[(1)]{\it Frame-level}: In this measure, if one pixel detects an anomaly 
then it is considered to be abnormal.
\item[(2)]{\it Pixel-level}: If at least 40 percent of anomaly ground truth pixels 
are covered by pixels that are detected by the algorithm, 
then the frame is considered to show an anomaly.
\end{itemize}

\subsection{Qualitative and Quantitative Results} 

\begin{figure}[t]
\center
  \includegraphics[width=1\linewidth]{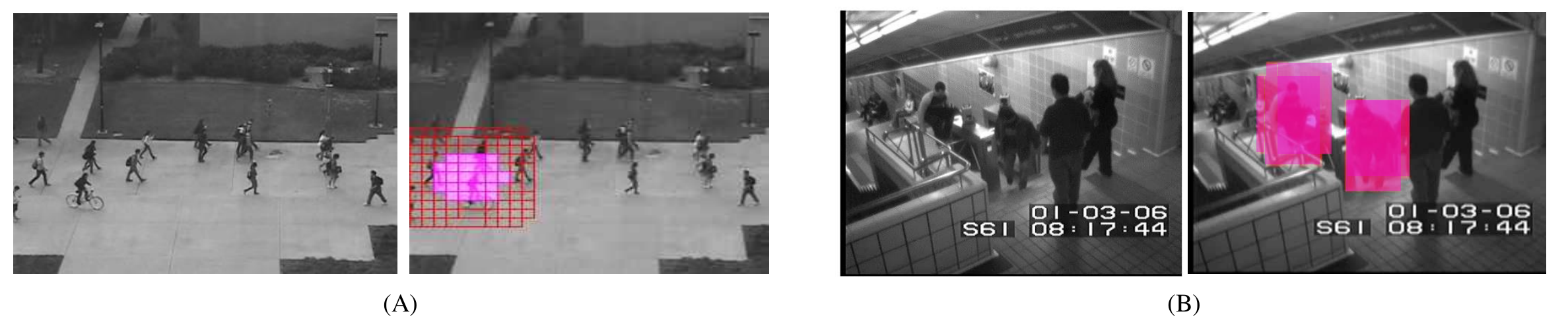}
   \caption
   {
   Output of the proposed method on Ped2 UCSD and Subway dataset. 
   {\it A-left and B-left:} Original frames. {\it A-Right and B-Right:} Anomaly regions  
   are indicated by red.  
   }
\label{fig:subway}
\vspace{2mm}
\end{figure}

Figure~\ref{fig:subway} illustrates the output of the proposed system on the samples of the
UCSD and Subway dataset. The proposed method detects and localizes
anomalies correctly in these samples. The main problem of an anomaly detection 
system is a high rate of false-positives. 

\begin{figure}[t]
\center
  \includegraphics[width=1\linewidth]{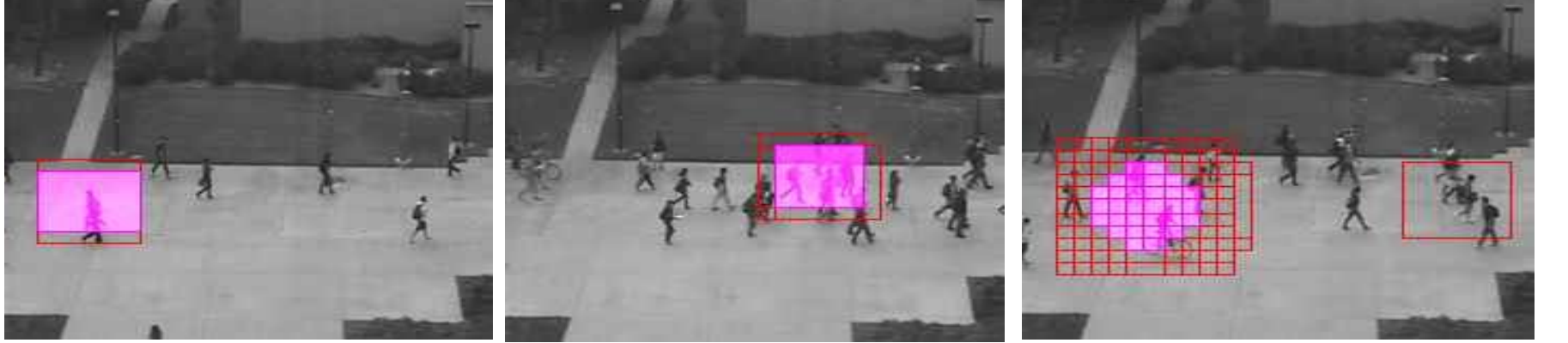}
\vspace{-6mm}
   \caption
   {
   Some examples of false-positives in our system. {\it Left:} 
   A pedestrian walking in opposite direction to other people. 
   {\it Middle:} A crowded region is wrongly detected as being an anomaly.
   {\it Right:} People walking in different directions.
   }
\label{fig:error}
\end{figure}

Figure~\ref{fig:error} shows regions which are wrongly 
detected as being an anomaly using our method.  
Actually, false-positives occur in two situations: too crowded scenes, and 
when people walk in different directions. 
Since walking in opposite direction of other pedestrians is not observed in 
the training video, this action is also considered as being abnormal using our algorithm.

\begin{figure}[t]
\center
  \includegraphics[width=1\linewidth]{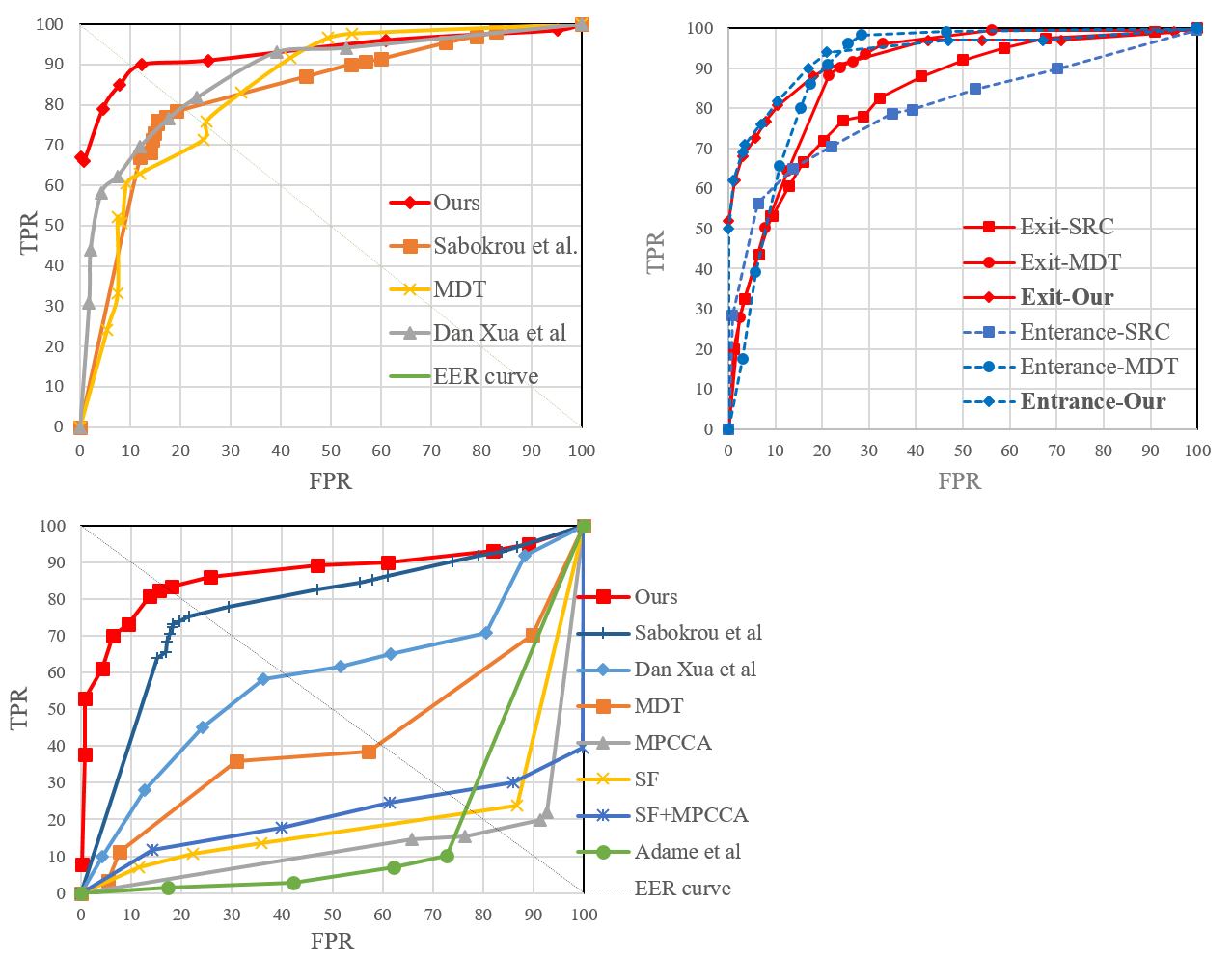}
   \caption
   {
   ROC comparison with state-of-the-art methods. {\it Upper left:} Frame-level of UCSD Ped2.  
   {\it Bottom left:} Pixel-level of UCSD Ped2. {\it Upper right:} Subway dataset.
   }
\label{fig:roc1}
\end{figure}

Frame-level and pixel-level ROCs of the proposed method in 
comparison to state-of-the-art methods are provided in Figure~\ref{fig:roc1}; left and 
middle for frame-level and pixel-level EER on UCSD Ped2 dataset, respectively.  
The ROCs show that the proposed method outperforms the other considered 
methods in the UCSD dataset.

\vspace{-2mm}

Table~\ref{tab:EER2} compares the frame-level and pixel-level EER of our method and 
other state-of-the-art methods. Our frame-level EER  is 
11$\%$, where the best result in general is 10$\%$, achieved by Tan Xiao et al.~\cite{xiao2015}. 
We outperform all other considered methods except~\cite{xiao2015}. 
On the other hand, the pixel-level EER of the proposed 
approach is 15\%, where the next best result is 17$\%$. As a result,
our method achieved a better performance than any other state-of-the-art method 
in the pixel-level EER metric by $2\%$.

The frame-level ROC of the Subway dataset is shown in 
Figure~\ref{fig:roc1} (right). In this dataset, we evaluate our method in both the entrance 
and exit scenes. The ROC confirms that our 
method has a better performance than MDT~\cite{Li2014} and SRC~\cite{cong2011} methods. 
We also discuss the comparison of AUC and EER in this dataset in Table~\ref{tab:subway}. 

For the exit scene, we outperform the other considered methods
in respect to both AUC and EER measures; we outperform 
by 0.5$\%$ and 0.4$\%$ in AUC and EER, respectively. For the 
entrance scenes, the AUC of the proposed method achieves 
better results compared to all other methods by 0.4$\%$. The proposed 
method gains better outcomes in terms of EER for all methods 
except Saligrama et al.~\cite{saligrama2012}; they achieve better results by 0.3$\%$. 

\begin{table}[t]
\caption{EER for frame and pixel level comparisons on Ped2; we only list first author
in this table for reasons of available space}
\vspace{-6mm}
\begin{center}
{\small
\begin{tabular}{lccccc}
\hline
Method   &	Frame-level  &  Pixel-level & Method   &	Frame-level  &  Pixel-level  \\
\hline\hline
IBC~\cite{Boiman2007} & 13\% &  26\% & Reddy~\cite{reddy2011}& 20\% &  --- \\
Adam~\cite{adam2008}& 42\% &  76\% &Bertini~\cite{{Bertini2012}}& 30\% &  ---\\
SF~\cite{mehran2009} & 42\% &  80\% &Saligrama~\cite{saligrama2012} &18\%\\
MPCCA~\cite{kim2009} & 30\% &  71\% &Dan Xu~\cite{xu2015} &20\% &42\%\\
MPCCA+SF~\cite{Mahadevan2010} & 36\% &  72\% &Li~\cite{Li2014}& 18.5\% & 29.9\%\\
 Zaharescu~\cite{Zaharescu2010} &17\% &30\%& Tan Xiao~\cite{xiao2015} & \textbf{10\%} &17\%\\  
MDT~\cite{Mahadevan2010} & 24\% & 54\% & Sabokrou~\cite{sabokrou2015}  &  19\% &24\% \\ \hline \hline
Ours & 11\%  & \textbf{15\%}  & &    \\
\hline
\end{tabular}
}
\end{center}
\label{tab:EER2}
\end{table}

\begin{table}[ht] 
\caption{AUC-EER comparison on Subway dataset}
\begin{center}
\begin{tabular}{lcccc}
\hline
Method & SRC~\cite{cong2011} & MDT~\cite{Mahadevan2010} & Saligrama et al.~\cite{saligrama2012} &Ours\\\hline\hline
Exit &80.2/26.4~~&89.7/16.4~~&88.4/17.9~~&\textbf{90.2/16}\\
Entrance &83.3/24.4 & 90.8/16.7  &  --/--  & \textbf{90.4}/17\\
\hline
\end{tabular}
\end{center}
\label{tab:subway}
\vspace{-4mm}
\end{table}

\begin{table}[b!] 
\vspace{-4mm}
\caption{Details of run-time (second/frame)}
\vspace{-6mm}
\begin{center}
\begin{tabular}{lcccc}
\hline
 & \quad Pre-processing \quad& \quad Representation \quad& \quad Classifying \quad & \quad Total \quad \\
 \hline\hline
\textbf{Time (in sec)} &0.0010&0.0016 &0.0001 & 0.0027\\
\hline
\end{tabular}
\end{center}
\label{tab:det_runtime}
\end{table}

\subsection{Run-time Analysis}

For processing a frame, three steps need to be performed:
Some pre-processing such as resizing the frames and constructing the input of the FCN, and 
representing the input by the FCN are considered as the first and second step, respectively. 
In the final step, the 
regional descriptors must be checked by a Gaussian classifier. 

With respect to these three steps, run-time details of our proposed method for processing 
a single frame are provided in Table~\ref{tab:det_runtime}. 
The total time for detecting an anomaly in a frame is 
$\approx$0.0027 sec. Thus, we achieve 370 fps, and this is much faster 
than any of the other considered state-of-the-art methods.  

Table~\ref{tab:runtime} 
shows the speed of our method in comparison to other approaches.
There are some key points which make our system fast. The proposed 
method benefits from fully convolutional neural networks. These types of 
networks perform feature extraction and localization concurrently. This 
property leads to less computations. 

\begin{table}[ht] 
\caption{Run-time comparison on Ped2 (in sec)}
\begin{center}
\vspace{-4mm}
{\small
\begin{tabular}{ccccccc}
\hline
\textbf{Method} &\shortstack{IBC~ \\ \cite{Boiman2007}}  & \shortstack{MDT\\~\cite{Mahadevan2010}} & \shortstack{~Roshtkhari et al. \\~\cite{Roshtkhari2013}}& \shortstack{Li et al.\\~\cite{Li2014}}& \shortstack{ Xiao et al.\\~\cite{xiao2015}}&Ours\\
\hline\hline
\textbf{Run-time} & 66 &23& 0.18 & 0.80&0.29&\textbf{$\approx$0.0027}\\
\hline
\end{tabular}
}
\end{center}
\label{tab:runtime}
\end{table}

Furthermore, by combining six frames into a three-channel input, 
we process a cubic patch of video frames at just one forward-pass. 
As mentioned before, for detecting abnormal regions, we only process 
two convolutional layers, and for some regions we classify them 
using a sparse auto-encoder. Processing these shallow layers 
results in reduced computations. Considering these tricks, besides processing 
fully convolutional networks in parallel, results in 
faster processing for our system compared to other methods.

\section{Conclusions}
\label{sec:c}

This paper presents a new FCN architecture for generating and describing 
abnormal regions for videos. By using the strength of FCN architecture 
for patch-wise operations on input data, the generated regional features are 
context-free. Furthermore, the proposed FCN is a combination of a 
pre-trained CNN (an AlexNet version) and a new convolutional layer 
where kernels are trained with respect to the chosen training video.
This final convolutional layer of the proposed FCN needs to be trained.
The proposed approach outperforms existing methods in processing speed. 
Besides, it is a solution for overcoming limitations in training samples 
used for learning a complete CNN. This method enables us 
to run a deep learning-based method at 
a speed of about 370 fps. Altogether, the proposed method is both
fast and accurate for anomaly detection in video data.

\section*{Acknowledgement}
This research was in part supported by a grant from IPM. (No. CS1396-5-01)

\end{document}